\documentclass[graybox]{svmult}

\usepackage{type1cm}        
\usepackage{makeidx}         
\usepackage{graphicx}        
\graphicspath{ {../images/} }

\usepackage{multicol}        
\usepackage[bottom]{footmisc}

\usepackage[hyphens]{url}

\usepackage{hyperref}
\usepackage{algorithm}
\usepackage{algpseudocode}

\usepackage{booktabs}

\usepackage{array}
\newcolumntype{P}[1]{>{\raggedright\arraybackslash}p{#1}}

\usepackage{newtxtext}       %
\usepackage{newtxmath}       

\makeindex             

\begin{document}

\title*{A Review of Privacy-preserving Federated Learning for the Internet-of-Things}
\author{Christopher Briggs, Zhong Fan and Peter Andras}
\institute{Christopher Briggs \at Keele University, Staffordshire, UK, \email{c.briggs@keele.ac.uk}
\and Zhong Fan \at Keele University, Staffordshire, UK \email{z.fan@keele.ac.uk}
\and Peter Andras \at Keele University, Staffordshire, UK \email{p.andras@keele.ac.uk}}
%
%
\maketitle

\abstract*{The Internet-of-Things (IoT) generates vast quantities of data, much of it attributable to individuals' activity and behaviour.  Gathering personal data and performing machine learning tasks on this data in a central location presents a significant privacy risk to individuals as well as challenges with communicating this data to the cloud. However, analytics based on machine learning and in particular deep learning benefit greatly from large amounts of data to develop high-performance predictive models. This work reviews federated learning as an approach for performing machine learning on distributed data with the goal of protecting the privacy of user-generated data as well as reducing communication costs associated with data transfer. We survey a wide variety of papers covering communication-efficiency, client heterogeneity and privacy preserving methods that are crucial for federated learning in the context of the IoT. Throughout this review, we identify the strengths and weaknesses of different methods applied to federated learning and finally, we outline future directions for privacy preserving federated learning research, particularly focusing on IoT applications.}

\abstract{The Internet-of-Things (IoT) generates vast quantities of data, much of it attributable to individuals' activity and behaviour. Gathering personal data and performing machine learning tasks on this  data in a central location presents a significant privacy risk to individuals as well as challenges with communicating this data to the cloud. However, analytics based on machine learning and in particular deep learning benefit greatly from large amounts of data to develop high-performance predictive models. This work reviews federated learning as an approach for performing machine learning on distributed data with the goal of protecting the privacy of user-generated data as well as reducing communication costs associated with data transfer. We survey a wide variety of papers covering communication-efficiency, client heterogeneity and privacy preserving methods that are crucial for federated learning in the context of the IoT. Throughout this review, we identify the strengths and weaknesses of different methods applied to federated learning and finally, we outline future directions for privacy preserving federated learning research, particularly focusing on IoT applications.}

\section{Introduction}
The Internet-of-Things (IoT) is represented by network-connected machines, often small embedded computers that provide physical objects with digital capabilities such as identification, inventory tracking and sensing \& actuator control. Mobile devices such as smartphones also represent a facet of the IoT, often used as a sensing device as well as to control and monitor other IoT devices. The applications that drive analytical insights in the IoT are often powered by machine learning and deep learning.

Gartner \cite{Gartner:2018wt} predicts that 25 billion IoT devices will be in use by 2021, forecasting a bright future for IoT applications. However this poses a challenge for traditional cloud-based IoT computing. The volume, velocity and variety of data streaming from billions of these devices requires vast amounts of bandwidth which can become extremely cost prohibitive. Additionally, many IoT applications require very low-latency or near real-time analytics and decision making capabilities. The round-trip delay from devices to the cloud and back again is unacceptable for such applications. Finally, transmitting sensitive data collected by IoT devices to the cloud poses security and privacy concerns. Edge computing, and more recently, fog computing \cite{Bonomi:2012kna} have been proposed as a solution to these problems.

Edge computing (and its variants: mobile edge computing, multi-access edge computing) restrict analytics processing to the edge of the network – on devices attached to, or very close to the perception layer \cite{Ai:2018kj}. However storage and compute power may be severely limited and coordination between multiple devices may be non-existent in the edge computing paradigm. Fog computing \cite{Bonomi:2012kna} offers an alternative to cloud computing or edge computing alone for many analytics tasks but significantly increases the complexity of an IoT network. Fog computing is generally described as a continuum of compute, storage and networking capabilities to power applications and services in one or more tiers that bridge the gap between the cloud and the edge \cite{Bittencourt:2018cr, Anonymous:QblbN7U6}. Fog computing enables highly scalable, low-latency, geo-distributed applications, supporting location awareness and mobility \cite{Bonomi:2014is}. Despite rising interest in fog-based computing, much research is still focused on deployment of analytics applications (including deep learning applications) directly to edge devices.

Performing computationally expensive tasks such as training deep learning models on edge devices poses a challenge due to limited energy budgets and compute capabilities \cite{Li:2018hz}. In cloud environments, massively powerful and scalable servers making use of parallelisation are typically employed for deep learning tasks \cite{BenNun:2019if}. In edge computing environments, alternative methods for distributing training are required. Additionally, as limited bandwidth is a key constraint in computing near/at the edge, the challenge of reducing network data transfer is also important. Federated learning \cite{mcmahan2017communication} has been proposed as a method for distributed machine learning, suitable for edge computing environments addresses many of the issues discussed above - namely, compute power, data transfer as well as privacy preservation.

This review provides a comprehensive survey of privacy preserving federated learning. We show how federated learning is ideally suited for data analytics in the IoT and review research addressing privacy concerns \cite{Abadi:2016gi}, bandwidth limitations \cite{Shokri:2015dr}, and power/compute limitations \cite{Wang:2019kq}. The rest of this review is organised as follows. Section \ref{sec:distributed-machine-learning} provides an introduction to preliminary work on distributed machine learning and its influence on federated learning literature. Section \ref{sec:federated-learning} describes federated learning in detail and outlines the major contributions to federated learning research including methods for reducing communication. Following this, section \ref{sec:privacy-preservation} gives an overview of privacy in data analysis and methods for preserving the privacy of an individual's data. Section \ref{sec:privacy-preservation-in-federated-learning} follows with an analysis of privacy preserving methods as applied to federated learning to protect latent data. Finally section \ref{sec:challenges-and-future-directions} discusses major outstanding challenges and future directions to apply federated learning to IoT applications and section \ref{sec:conclusion} presents concluding remarks.

\section{Distributed machine learning}
\label{sec:distributed-machine-learning}

Federated learning was preceded by much work in distributed machine learning in the data-centre \cite{Dean:2012wx, BenNun:2019if, Li:tt}. This section gives a brief history of distributed machine learning, paying particular attention to distributed deep learning training via stochastic gradient descent (SGD). Deep learning is concerned with machine learning problems based on artificial neural networks comprised of many layers and has been used with great success in the fields of computer vision, speech recognition and translation as well as many other areas \cite{2015Natur.521..436L}. In these fields, most other machine learning methods have been surpassed by deep learning methods due to the very complex functions they can compute which can both approximate training labels and generalise well to unseen samples. 

Deep neural networks (DNNs) are composed of multiple connected units (also known as neurons) organised into layers through which the training data flows \cite{Goodfellow-et-al-2016}. Each unit computes a weighted sum of its input values (including a bias term) composed with a non-linear activation function $g(\mathbf{W}^\top \mathbf{X} + \mathbf{b})$ and returns the result to the next connected layer. Passing data through the network and performing a prediction is known as the forward pass. To train the network, a backward pass operation is specified to compute updates to the weights and biases in order to better approximate the labels associated with the training data. An algorithm known as backpropagation \cite{RUMELHART:1986to} is used to propagate the error back through each layer of the network by calculating gradients of the weights and biases with respect to the error.

DNNs perform best when trained on very large datasets and often incorporate millions if not billions of parameters to express weights between neurons (for example the AlexNet DNN achieved state-of-the-art performance on the ImageNet dataset in 2012 using 60 million parameters \cite{Krizhevsky:2017gx}). Both of these factors require large sums of memory and compute capabilities. To scale complex DNNs trained on lots of data requires concurrency across multiple CPUs or more commonly GPUs (most often in a local cluster). GPUs are optimised to perform matrix calculations and are well suited for the operations required to compute activations across a DNN. Concurrency can be achieved in a variety of ways as discussed below.

\subsection{Concurrency}
To train a large DNN efficiently across multiple nodes, the calculations required in the forward and backward passes need to parellelised. One method to achieve this is model parallelism which distributes collections of neurons among the available compute nodes \cite{Dean:2012wx}. Each node then only needs to compute the activations of its own neurons, however must communicate regularly with nodes computing on connected neurons. The calculations on all nodes must occur synchronously and therefore computation proceeds at the speed of the slowest node in the cluster. Another drawback of the model parallelism approach is that the current mini-batch must be copied to all nodes in the compute cluster, further increasing communication costs within the cluster.

A second method resolves some of the issues of excessive communication between nodes by distributing one or more layers on each node. This ensures that that each worker node only needs to communicate with the one other node (a different node depending whether the computation is part of the forward pass or the backward pass) \cite{BenNun:2019if}. However, this method still requires that data in the mini-batch be copied to all nodes in the cluster.

The final method to achieve parallelism in training a large DNN is termed data parallelism. This method partitions the training dataset and copies the subsets to each compute node in the cluster. Each node computes forward and backward passes over the same model but using mini-batches drawn from its own subset of the training data. The results of the weight updates are then reduced on each iteration via MapReduce or more commonly today, via message passing interface (MPI) \cite{BenNun:2019if}. Data parallelism is particularly effective as most operations over mini-batches in SGD are independent. Therefore scaling the problem via sharding the data to many nodes is relatively simple compared to the methods mentioned above. This method solves the issue of training with large amounts of data but requires that the model (and its parameters) fit in memory on each node.

Hybrid parallelism combines two or all three of the concurrency schemes mentioned above to mitigate the drawbacks associated with each and best support parallelism on the underlying hardware. DistBelief \cite{Dean:2012wx} achieves this by distributing the data, network layers, and neurons within the same layer among the available compute nodes, making use of all three concurrency schemes. Similarly, Project Adam \cite{Chilimbi:vp} employs all three concurrency schemes but much more efficiently than DistBelief (using significantly fewer nodes to achieve high accuracy on the ImageNet\footnote{http://www.image-net.org/} 22k data set)

\subsection{Model consistency}
Model consistency refers to the state of a model when trained in a distributed manner \cite{BenNun:2019if} - a consistent model should reflect the same parameter values among compute nodes prior to each training iteration (or set of training iterations, sometimes referred to as a communication round). In order to maintain model consistency, individual compute nodes need to write updates to a global parameter server \cite{Li:tt}. The parameter server performs some form of aggregation on the updates to synchronise a global model and the parameters (for example, weights in a neural network) are then shared with the individual compute nodes for the next iteration/round of training.

There are several broad methods by which to train, update and share a distributed deep learning model. Synchronous updates occur when the parameter server waits for all compute nodes to return parameters for aggregation. This method provides high consistency between iterations/rounds of training as each node always receives up-to-date parameters but is not hardware performant due to delays caused by the slowest communicating node. For example, a set of parameters $w_t$ at time $t$ is shared among $n_c$ compute nodes. The compute nodes each perform some number of forward and backward passes over the data available to them and compute the parameter gradients $\Delta w_c$. These gradients are communicated to the parameter server, which in turn averages the gradients from all workers and then updates the parameters for time $t+1$:

\begin{equation}
	\label{eq:synchronous-parameter-update}
	\begin{aligned}
	\Delta w_t &= \frac{1}{n_c} \sum_{c=1}^{n_c} \Delta w_c. \\
	w_{t+1} &= w_t - \eta \Delta w_t.
	\end{aligned}
\end{equation}

Asynchronous updates occur when the parameter server shares the latest parameters without waiting for all nodes to return parameter updates. This reduces model consistency as parameters can be overwritten and become stale due to slow communicating nodes. This method is hardware performant however as optimisation can proceed without waiting for all nodes to send parameter updates. The HOGWILD! algorithm \cite{Recht:2011wo} takes advantage of sparsity within the parameter update matrix to asynchronously update gradients in shared memory resulting in faster convergence. Downpour SGD \cite{Dean:2012wx} describes asynchronous updates as an additional mechanism to add stochasticity to the optimisation process resulting in greater prediction performance. 

In order to improve consistency using hardware performant asynchronous updates, the concept of parameter `staleness' has been tackled by several works. The stale synchronous parallel (SSP) model \cite{Ho:2013wd} synchronises the global model once a maximum staleness threshold has been reached but still allows workers to compute updates on stale values between global model syncs. The impact of staleness in asynchronous SGD can also be mitigated by adapting the learning rate as a function of the parameter staleness \cite{2016arXiv160104033O, 2015arXiv151105950Z}. As an example a worker pushes an update at $t = j$ to the parameter server at $t = i$. The parameters in the global model at $t = i$ are the most up-to-date available. To prevent a stale parameter update from occurring, a staleness parameter $\tau_k$ for the $k$-th parameter is calculated as $\tau_k = i - j$. The learning rate used in \autoref{eq:synchronous-parameter-update} is modified as:

\begin{equation}
	\eta_k = 
	\begin{cases}
		\eta/\tau_k & \text{if } \tau_k \neq 0\\
		\eta & \text{otherwise}
	\end{cases}.
\end{equation}

\subsection{Centralised vs decentralised learning}
Centralised distribution of the model updates requires a parameter server (which may be a single machine or sharded across multiple machines as in \cite{Dean:2012wx}). The global model tracks the averaged parameters aggregated from all the compute nodes that perform training (see \autoref{eq:synchronous-parameter-update}). The downside to this distribution method is the high communication cost between compute nodes and the parameter server. Multiple shards can relieve this bottleneck to some extent, such that different workers read and write parameter updates to specific shards \cite{Chilimbi:vp, Dean:2012wx}. 

Heterogeneity of worker resources is handled well in centralised distribution models. Distributed compute nodes introduce varying amounts of latency (especially when distributed geographically as in \cite{Hsieh:tv}), yet training can proceed via asynchronous, or more efficiently, stale-synchronous methods \cite{Jiang:2017cs}. Heterogeneity is an inherent feature of \hyperref[sec:federated-learning]{federated learning}.

Decentralised distribution of DNN training does not rely on a parameter server to aggregate updates from workers but instead allows workers to communicate with one another, resulting in each worker performing aggregation on data from the parameters it receives. Gossip algorithms that share updates between a fixed number of neighbouring nodes have been applied to distributed SGD \cite{2018arXiv180305880D, 2016arXiv161104581J, SundharRam:2009fa} in order to efficiently communicate/aggregate updates between all nodes in an exponential fashion similar to how disease is spread during an epidemic.

Communication can be avoided completely during training, resulting in many individual models represented by very different parameters. These models can be combined (as an ensemble \cite{2015Natur.521..436L}), however averaging the predictions from many models can slow down inference on new data. To tackle this, a process known as knowledge distillation can be used to train a single DNN (known as the mimic network) to emulate the predictions of an ensemble model \cite{Ba:2014tr, Chebotar:2016bi, 2015arXiv150302531H}. Unlabelled data is passed through the ensemble network to obtain labels on which the mimic network can be trained.
\section{Federated learning}
\label{sec:federated-learning}

\subsection{Overview}
Federated learning extends the idea of distributed machine learning, making use of data parallelism. However, rather than randomly partitioning a centralised dataset to many compute nodes, training occurs in the user domain on distributed data owned by the individual users (often referred to as clients) \cite{mcmahan2017communication}. The consequence of this is that user data is never shared directly with a third party orchestrating the training procedure. This greatly benefits users where the data might be considered sensitive. Where data needs to be observed (for example, during the training operation), processing is handled on the device where the data resides (for example a smartphone). Once a round of training is completed on the device, the model parameters are communicated to an aggregating server, such as a parameter server provided by a third party. Although the training data itself is never disclosed to the third-party, it is a reasonable concern that something about an individual's training data might be inferred by the parameter updates; this is discussed further in \autoref{sec:privacy-preservation-in-federated-learning}.

Federated learning is vastly more distributed than traditional approaches for training machine learning models via data parallelism. Some of the key differences are \cite{mcmahan2017communication}:

\begin{enumerate}
	\item \textbf{Many contributing clients} - federated learning needs to be scalable to many millions of clients.
	\item \textbf{Varying quantity of data owned by each user} - some clients may train on only a few samples; others may have thousands.
	\item \textbf{Often very different data distributions between users} - user data is highly personal to individuals and therefore the model trained by each client represents non-IID (independent, identically distributed) data.
	\item \textbf{High latency between clients and aggregating service} - updates are commonly communicated via the internet introducing significant latency between communication rounds.
	\item \textbf{Unstable communication between clients and aggregating service} - client devices are likely to become unavailable during training due to their mobility, battery life, or other reasons.
\end{enumerate}

These distinguishing features of federated learning pose challenges above and beyond standard distributed learning.

Although this review focuses on deep learning in particular, many other ML algorithms can be trained via federated learning. Any ML algorithm designed to minimise an objective function of the form:

\begin{equation}
    \label{eq:risk-minimisation}
    \min_{w \in \mathbb{R}^d} \frac{1}{m} \sum_{i=1}^{m} f_i(w).
\end{equation}

is well suited to training via many clients (for example linear regression and logistic regression). Some non-gradient based algorithms can also be trained in this way, such as principal component analysis and k-mean clustering \cite{Liang:vq}.

Federated optimisation was first suggested as a new setting for vastly and unevenly distributed machine learning by Kone\u{c}n\'{y} et al. \cite{2016arXiv161002527K} in 2016. In their work, the authors first describe the nature of the federated setting (non-IID data, varying quantity of data per client etc). Additionally, the authors test a simple application of distributed gradient descent against a federated modification of SVRG (a variance reducing variant of SGD \cite{Johnson:2013:ASG:2999611.2999647}) over distributed data. Federated SVRG calculates gradients and performs parameter updates on each of $K$ nodes over the available data on each node and obtains a weighted average of the parameters from all clients. The performance of these algorithms are verified on a logistic regression language model using Google+ data to determine whether a post will receive at least one comment. As logistic regression is a convex problem, the algorithms can be benchmarked against a known optimum. Federated SVRG is shown to outperform gradient descent by converging to the optimum within 30 rounds of communication.

\begin{algorithm}
	\caption{Federated Averaging (\textsc{FedAvg}) algorithm. $C$ is the fraction of clients selected to participate in each communication round. The $K$ clients are indexed by $k$; $B$ is the local mini-batch size, $\mathcal{P}_k$ is the dataset available to client $k$, $E$ is the number of local epochs, and $\eta$ is the learning rate}
	\label{alg:federatedAveraging}
	\begin{algorithmic}[1]
	\Procedure{FedAvg}{}\Comment{Run on server}
		\State Initialise $w_0$
		\For {each round $t = 1,2,...$}
			\State $m \gets \max{(C \cdot K, 1)}$
			\State $S_t \gets$ (random set of $m$ clients)
			\For {each client $k \in S_t$}\Comment{In parallel}
				\State $w_{t+1}^k \gets$ \textsc{ClientUpdate}($k, w_t$)
			\EndFor
			\State $w_{t+1} \gets \sum_{k=1}^K \frac{n_k}{n} w_{t+1}^k$
		\EndFor
	\EndProcedure
	\Statex
	\Procedure{ClientUpdate}{$k, w$}\Comment{Run on client $k$}
	\State $\mathcal{B}\gets$ (Split $\mathcal{P}_k$ into mini-batches of size $B$)
	\For {each local epoch $i$ from 1 to $E$}
		\For {batch $b \in \mathcal{B}$}
			\State $w \gets w - \eta \nabla \mathcal{L}(w;b)$
		\EndFor
	\EndFor
	\State return $w$ to server
	\EndProcedure
	\end{algorithmic}
\end{algorithm}
	
Federated learning (as described in \cite{mcmahan2017communication} simplifies the federated SVRG approach in \cite{2016arXiv161002527K} by modifying SGD for the federated setting. McMahan et al. \cite{mcmahan2017communication} provide two distributed SGD scenarios for their experiments: FedSGD and FedAvg. FedSGD performs a single step of gradient descent on all the clients and averages the gradients on the server. The FedAvg algorithm (shown in algorithm \autoref{alg:federatedAveraging}) randomly selects a fraction of the clients to participate in each round of training. Each client $k$ computes the gradients on the current state of the global model $w_t$ and updates the parameters $w_{t+1}^k$ in the standard fashion in gradient descent:

\begin{equation}
	\forall k, w_{t+1}^{k} \leftarrow w_t - \eta \nabla f(w_t).
\end{equation}

All clients communicate their updates to the aggregating server, which then calculates a weighted average of the contributions from each client to update the global model:

\begin{equation}
	w_{t+1} \leftarrow \sum_{k=1}^{K} \frac{n_k}{n} w_{t+1}^k.
\end{equation}

\begin{figure*}[h]
    \centering
    \includegraphics[width=\linewidth]{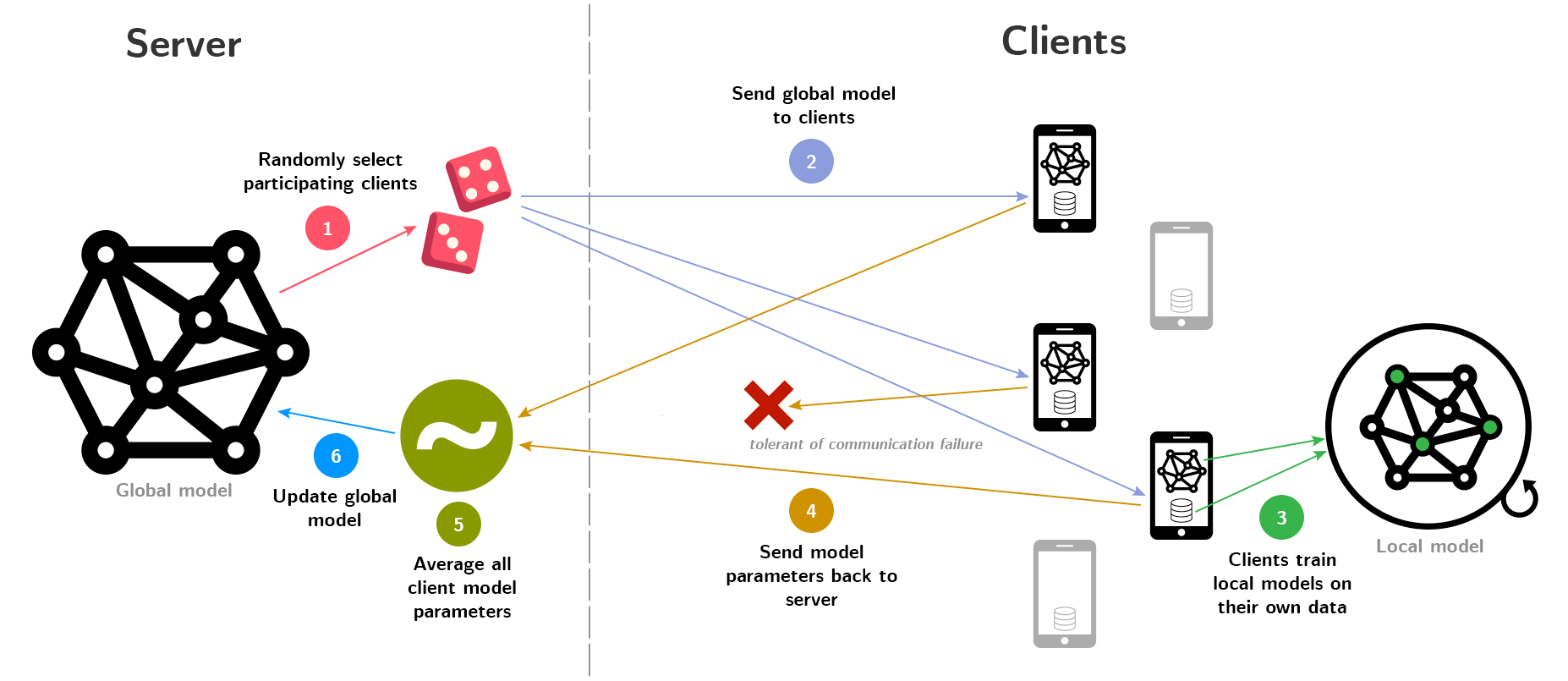}
    \caption{Schematic diagram showing how communication proceeds between the aggregating server and individual clients according to the FedAvg protocol. This procedure is iterated until the model converges or the model reaches some desired target metric (e.g. elapsed time, accuracy)}
    \label{fig:federated-learning-schematic}
\end{figure*}

Here, $n_k/n$ is the fraction of data available to the client compared to the available data to all participating clients. Clients can perform one or multiple steps of gradient descent before sending weight updates as orchestrated by the federated algorithm. A diagram describing how federated learning proceeds in the FedAvg scenario is provided in \autoref{fig:federated-learning-schematic}

Centralised machine learning (and distributed learning in the data center) benefits from training under the assumption that data can be shuffled and is independent and identically distributed (IID). This assumption is generally invalid in federated learning as the training data is decentralised with significantly different distributions and number of samples between participating clients. Training using non-IID data has been shown to converge much more slowly than IID data in a federated learning setting using the MNIST\footnote{http://yann.lecun.com/exdb/mnist/} dataset (for handwritten digit recognition), distributed between clients after having been sorted by the target label \cite{mcmahan2017communication}. The overall accuracy achieved by a DNN trained via federated learning can be significantly reduced when trained on highly skewed non-IID data \cite{2018arXiv180600582Z}. Yue et al. \cite{2018arXiv180600582Z} show that accuracy can be improved by sharing a small subset of non-private data between all the clients in order to reduce the variance between weight updates of the clients involved in each communication round. The FedProx algorithm \cite{mlsys2020_176} encompasses FedAvg as a special case and adds a regularising term to the local optimisation objective. This has the effect of limiting the distance between the local model and global model during each communication round and stabilises training overall. Karimireddy et al \cite{karimireddy2019scaffold} takes a similar approach using SCAFFOLD by accounting for client drift (the estimated difference between the global and local model directions) and corrects for this in the model update step. This can be understood as a variance reduction method and significantly outperforms FedAvg by reducing the number of rounds of communication and improving the final model accuracy on highly skewed non-IID data.

\begin{table*}
\caption{A summary of important contributions to federated learning research}
\label{table:federated-learning-research}
\centering
\small
\begin{tabular}{c P{1in} c p{3in}}
\toprule
\textbf{Ref} & \textbf{Research focus} & \textbf{Year} & \textbf{Major contribution}\\
\midrule
\cite{2016arXiv161002527K} & Optimisation & 2016 & First description of federated optimisation and its application to a convex problem (logistic regression)\\
\midrule
\cite{mcmahan2017communication}         & Optimisation                         & 2016 & Description of federated averaging (FedAvg) algorithm to improve the performance of the global model and reduce communication between the clients and server\\
\midrule
\cite{Konecny:2016ts}         & Communication                         & 2016 & Methods for compressing weight updates and reducing the bandwidth required to perform federated learning\\
\midrule
\cite{2017arXiv170510467S}    & Multi-task FL        & 2017 & Application of multi-task learning in a federated setting and discussion of system challenges relevant to using federated learning on resource-constrained devices\\
\midrule
\cite{2018arXiv180700459B}    & FL attacks                         & 2018 & A demonstration of poisoning the shared global model in a federated learning setting\\
\midrule
\cite{2018arXiv180804866F}    & FL attacks     & 2018 & A method to recognise adversarial clients and combat model poisoning in a federated learning setting\\
\midrule
\cite{Hard:2018tda}           & Application                                          & 2018 & Application of federated learning in a commercial setting (next word keyborard prediction in Android Gboard)\\
\midrule
\cite{2018arXiv181005512L}    & Optimisation                           & 2018 & Application of per-coordinate averaging (based on Adam) to federated learning to achieve faster convergence (in fewer communication rounds)\\
\midrule
\cite{Liu:2018tv}             & Application                                   & 2018 & Applied federated learning to a healthcare application including further training after federated learning on client data (transfer learning)\\
\midrule
\cite{2018arXiv180408333N}    & Client selection                              & 2018 & A method of federated learning selecting clients with faster communication/greater resources to participate in each communication round achieving faster convergence\\
\midrule
\cite{Wang:2019kq}            & Communication & 2018 & Description of adaptive federated learning method suitable for deployment on resource-constrained devices to optimally learn a shared model while maintaining a fixed energy budget\\
\midrule
\cite{2018arXiv180600582Z}    & Non-IID          & 2018 & Characterisation of how non-IID data reduces the model performance of federated learning and method for improving model performance\\
\midrule
\cite{mlsys2020_176}    & Optimisation          & 2018 & Adds a tunable regularising term to FedAvg to stabilise training on skewed, non-IID data, limiting the influence of client models on the global model. \\
\midrule
\cite{2019arXiv190410120E}    & Multi-task FL & 2019 & Training pluralistic models that are tailored to subsets of clients that belong to the same timezones \\
\midrule
\cite{karimireddy2019scaffold}    & Optimisation & 2020 & Applies a variance reduction method for improving convergence speed on non-IID data compared to FedAvg \\
\bottomrule
\end{tabular}
\end{table*}

\subsection{Multi-task federated learning}

A different approach to federating optimisation over many nodes is proposed by Smith et al. \cite{2017arXiv170510467S}. In this work, each client's data distribution is modelled as a single task as part of a multi-task learning objective. In multi-task learning, all tasks are assumed to be similar and therefore each task can benefit from learning derived from all the other tasks.  On three different problems (based on human activity recognition and computer vision), the federated multi-task learning setting outperforms a centralised global setting and a strictly localised setting with lower average prediction errors. As part of this work \cite{2017arXiv170510467S}, the authors also show that federated multi-task learning is robust to nodes temporarily dropping out during learning and when communication is reduced (by simulating more iterations on the client per communication round). Eichner et al. \cite{2019arXiv190410120E} propose a pluralistic approach to tackle the issue of training only when devices are available (generally overnight for mobile phones). Multiple models are trained according to the timezone when the device is available and results in better language models targeted at each timezone. To specifically tackle the issue to model degredation  due to the presence of non-IID data \cite{2018arXiv180600582Z}, Sattler et al. \cite{Sattler:2019tb} propose splitting the shared model by determining the cosine similarity of updates from different clients during training. Similarly, Briggs et al. \cite{2020arXiv200411791B} use a hierarchical clustering algorithm to judge client update similarity to produce models tailored to clients with similarly-distributed data.

\subsection{Applied federated learning}
Federated learning is particularly well suited as a solution for distributed learning in the IoT setting. As such, federated learning research is flourishing in various applications associated with the IoT. Federated learning has been applied in robotics to aid multiple robots to share imitation learning strategies \cite{Liu:2020hr} and more generally for protecting privacy-sensitive robotics tasks \cite{Zhou:2018ez}. In mobile edge computing environments, federated learning has been demonstrated for predicting demand in edge deployed applications \cite{Fantacci:2020db} and for improving proactive edge content caching mechanisms \cite{Yu:2018fx}. For vehicular edge computing, Lu et al. \cite{Lu:2019jf} propose a framework to tackle issues of intermittent vehicle connectivity and an untrusted aggregating entity and Ye et al. \cite{Ye:2020em} propose a system using federated learning for intelligent connected vehicle image classification tasks. Energy demand in electrical vehicle charging networks has also been addressed with a federated learning strategy by Saputra et al. \cite{Saputra:2019ha}. For anomaly detection in IoT environments, federated learning has been applied to detect intrusions and attacks by Nguyen et al. \cite{Nguyen:2019by}. More novel applications of federated learning include learning to detect jamming in drone networks \cite{Mowla:2020fx}, predicting breaks in presence by users of virtual reality environments \cite{Chen:2020dj} and human activity recognition using wearable devices \cite{Sozinov:2018fk}.

For supervised problems, user data needs to be labelled by the user to be useful for training. This is demonstrated in \cite{Hard:2018tda} where a long short-term memory neural network (LSTM) is trained via many clients on words typed on a mobile keyboard to predict the next word. However, this data is clearly highly sensitive and should not be sent to a central server directly and would benefit from training via federated learning. The training data for this model is automatically labelled when the user types the next word. In cases where data is stored locally and already labelled such as medical health records, privacy is of great concern and even sharing of data between hospitals may be prohibited \cite{Miotto:2017km}. Federated learning can be applied in these settings to improve a shared global model that is more accurate than a model trained by each hospital separately. In \cite{Liu:2018tv}, electronic health records from 58 hospitals are used to train a simple neural network in a federated setting to predict patient mortality. The authors found that partially training the network using federated learning, followed by freezing the first layer and training only on the data available to each client resulted in better performing models for each hospital.

\subsection{Federated learning attacks}

Due to the nature of distributed client participation required for federated learning, the protocol is susceptible to adversarial attacks. Multiple works \cite{2018arXiv180700459B, Bhagoji:2019vz} present methods for poisoning the global model with an adversary acting as a client in the federated learning setting. The adversary constructs an update such that it survives the averaging procedure and heavily influences or replaces the global model. In this way, an adversary can poison the model to return predictions specified by the attacker given certain input features. Fung et al. \cite{2018arXiv180804866F} describe a method to defend against sybil-based adversarial attacks by measuring the similarity between client contributions during model averaging and filtering attacker's updates out. These kinds of attacks might be inadvertently mitigated against using some of the modifications to FedAvg outlined above (for example by FedProx \cite{mlsys2020_176} or SCAFFOLD \cite{karimireddy2019scaffold}) to limit the effect of individual client updates.

\subsection{Communication-efficient federated learning}
As highlighted above, distributed learning and federated learning in particular suffer from high latency between communication rounds. Additionally, given a sufficiently large DNN, the number of parameters that need to be communicated in each communication round from possibly many thousands of clients becomes problematic in relation to data transmission and traffic arriving at the aggregating server. There are several approaches to mitigating these issues as discussed in this subsection.

The simplest method to reduce bandwidth use in communicating model updates is simply to communicate less often. In \cite{mcmahan2017communication}, researchers experimented with the mini-batch size and number of epochs while training a convolutional neural network (CNN) on the MNIST dataset as well as an LSTM trained on the complete works of William Shakespeare\footnote{https://www.gutenberg.org/ebooks/100} to predict the next text character after some input of characters. Using FedAvg, the authors \cite{mcmahan2017communication} showed that increasing computation on the client between communication rounds significantly reduced the number of communication rounds required to converge to a threshold test accuracy compared to a single epoch trained on all available data on the client (a single iteration of gradient descent). The greatest reduction in communication rounds was achieved using a mini-batch size of 10 and 20 epochs on the client using the CNN model (34.8x) and mini-batch size of 10 and 5 epochs using the LSTM model (95.3x). As this method completely eliminates many of the communication rounds, it should be preferred over (or combined with) the compression methods discussed next.

As the network connection used to communicate between the clients and the aggregating server is generally asymmetric (download speed is generally significantly faster than upload speed), downloading the updated model from the aggregating server is less of a concern than uploading the updates from the clients. Despite this, compression methods exist to reduce the size of deep learning models themselves \cite{2015arXiv151000149H, Han:2015vz}.

The compression of the parameter updates on the client prior to transmission is important to reduce the size of the overall update but should still maintain a stable statistical mean when updates from all clients are averaged in the global shared model. The following compression methods are not specific to federated learning but have been experimented with in several works related to federated learning.

The individual weights that represent the parameters of a DNN are generally encoded using a 32-bit floating point number. Multiple works explore the effect of lossy compression of the weights (or gradients) to 16-bit \cite{2015arXiv150202551G}, 8-bit \cite{2015arXiv151104561D}, or even 1-bit \cite{Seide:vu} employing stochastic rounding to maintain the expected value. The results of these experiments show that as long as the quantisation error is carried forward between mini-batch computations, the overall accuracy of the model isn't significantly impacted. 

Another method to compress the weight matrix itself is to convert the matrix from a dense representation to a sparse one. This can be achieved by applying a random mask to the matrix and only communicating the resulting non-zeros values along with the seed used to generate the random mask \cite{Konecny:2016ts}. Using this approach combined with FedAvg on the CIFAR-10\footnote{https://www.cs.toronto.edu/~kriz/cifar.html} image recognition dataset, it has been shown \cite{Konecny:2016ts} that neither the rate of convergence nor the overall test accuracy is significantly impacted, even when only 6.25\% of the weights are transmitted during each communication round. Also described in \cite{Konecny:2016ts} is a matrix factorization method whereby the weight matrix is approximated by the product of a randomly generated matrix, $A$ and another matrix optimised during training, $B$. Only the matrix $B$ (plus the random seed to generate A) needs to be transmitted in each communication round. The authors show however that this method performs significantly worse than the random mask method as the compression ratio is increased.

Shokri \& Shmatikov \cite{Shokri:2015dr} propose an alternative sparsification method implemented in their ``Selective SGD'' (SSGD) procedure. This method transfers only a fraction of randomly selected weights to each client from the global shared model and only shares a fraction of updated weights back to the aggregating service. The updated weights selected to be communicated are determined by either weight size (largest unsigned magnitude updates) or a random subset of values above a certain threshold. The authors \cite{Shokri:2015dr} show that a CNN trained on the MNIST and Street View House Numbers (SVHN)\footnote{http://ufldl.stanford.edu/housenumbers/} datasets can achieve similar levels of accuracy sharing only 10\% of the updated weights and only a slight drop (1-2\%) in accuracy by sharing only 1\% of the updated weights. The paper also shows that the greater the number of users participating in SSGD, the greater the overall accuracy. 

Hardy et al. \cite{Hardy:hu} take a similar approach to selective SGD but select the largest gradients in each layer rather than the whole weight matrix to better reflect changes throughout the DNN. Their algorithm, ``AdaComp'' also uses an adaptive learning rate per parameter based on the staleness of the parameter to improve overall test accuracy on the MNIST dataset using a CNN. Most recently, Lin et al. \cite{2017arXiv171201887L} apply gradient sparsification along with gradient clipping and momentum correction during training to reduce communication bandwidth by 270x and 600x without a significant drop in prediction performance on various ML problems in computer vision, language modelling and speech recognition.

Leroy et al. \cite{2018arXiv181005512L} experiment with using moment-based averaging inspired by the Adam optimisation procedure, in place of a standard weighted global average in FedAvg. The authors train a CNN to detect a ``wake word'' (similar to ``Hey Siri'' to initialise the Siri program on iOS devices). The moment-based averaging method acheives a target recall of almost 95\% within 100 rounds of communication over ~1400 clients compared to only 30\% using global averaging.

By selecting clients based on client resource constraints in a mobile edge computing environment, Nishio \& Yonetani \cite{2018arXiv180408333N} show that federated learning can be sped up considerably. As federated learning proceeds in a synchronous fashion, the slowest communicating node is a limiting factor in the speed at which training can progress. In this work \cite{2018arXiv180408333N}, target accuracies on the CIFAR-10 and Fashion-MNIST\footnote{https://www.kaggle.com/zalando-research/fashionmnist} datasets are achieved in significantly less time than by using the FedAvg algorithm in \cite{mcmahan2017communication}. In a similar vein, Wang et al. \cite{Wang:2019kq}, aim to take into account client resources during federated learning training. In this work an algorithm is designed to control the tradeoff between local gradient updates and global averaging in order to optimally minimise the loss function under a fixed energy budget - an important problem for federated learning in the IoT (especially for battery-powered devices).
\section{Privacy preservation}
\label{sec:privacy-preservation}

Data collection for the purpose of learning something about a population (for example in machine learning to discover a function for mapping the data to target labels) can expose sensitive information about individual users. In machine learning, this is often not the primary concern of the developer or researcher creating the model, yet is extremely important for circumstances where personally sensitive data is collected and disseminated in some form (e.g. via a trained model). Privacy has become even more important in the age of big data (data which is characterised by its large volume, variety and velocity \cite{Gandomi:2015hh}). As businesses gather increasing amounts of data about users, the risk of privacy breaches via controlled data releases grows. 

This review focuses on the protection of an individual's privacy via controlled data releases (such as from personal data used to train a machine learning model) and does not consider privacy breaches via hacking and theft which is a separate issue related to data security.

Privacy is upheld as a human right in many countries via Article 12 of the Universal Declaration of Human Rights \cite{Anonymous:2015uv}, Article 17 of the International Covenant on civil and political rights \cite{Anonymous:1966uz}, and Article 8 of the European Convention on Human Rights \cite{Anonymous:1950wd}. In Europe rigorous legislation with respect to data protection via the General Data Protection Regulation \cite{eu:gdpr} safeguards data privacy such that users are given the facts about how and what data is collected about them and how it used and by whom. Despite these rights and regulations, data privacy is difficult to maintain and breaches of privacy via controlled data releases occur often.

Privacy can be preserved in a number of ways, yet it is important to maintain a balance between the level of privacy and utility of the data (along with some consideration for the computational complexity required to preserve privacy). A privacy mechanism augments the original data in order to prevent a breach of personal privacy (i.e. an individual should not be able to be recognised in the data). For example, a privacy mechanism might use noise to augment the result of a query on the data \cite{Dwork:2006dw}. Adding too much noise to a result might render it meaningless and adding too little noise might leak sensitive information. The privacy/utility tradeoff is a primary concern of the privacy mechanisms to be discussed in the next subsection.


\subsection{Privacy preserving methods}
The privacy preserving methods discussed in this section can be described as either \emph{suppressive} or \emph{perturbative} \cite{Fung:2010ena}. Suppressive methods include removal of attributes in the data, restricting queries via the privacy mechanism, aggregation/generalisation of data attributes and returning a sampled version of the original data. Perturbative methods include noise addition to the result of a query on the data or rounding of values in the dataset.

\subsubsection{Anonymisation}
Anonymisation or de-identification is achieved by removing any information that might identify an individual within a dataset. Ad-hoc anonymisation might reasonably remove names, addresses, phone numbers etc and replace each user's record(s) with a pseudonym value to act as an identifier under the assumption that individuals cannot be identified within the altered dataset. However, this leaves the data open to privacy attacks known as linkage attacks \cite{Fung:2010ena}. In the presence of auxiliary information, linkage attacks allow an adversary to re-identify individuals in the otherwise anonymous dataset. 

Several famous examples of such linkage attacks exist. An MIT graduate, Latanya Sweeney, purchased voter registration records for the town of Cambridge, Massachusetts and was able to use combinations of attributes (ZIP code, gender and date of birth) known as a \emph{quasi-identifier} to identify the then governor of Massachusetts, William Weld. When combined with state-released anonymised medical records, Sweeney was able to identify his medical information from the data release \cite{Greely:2007jc}. As part of a machine learning competition known as the Netflix Prize\footnote{https://www.netflixprize.com/index.html} launched in 2006, Netflix released a random sample of pseudo-anonymised movie rating data. Narayanan \& Shmatikov \cite{Narayanan:2008iu} were able to show that using relatively few publicly published ratings by IMDb\footnote{https://www.imdb.com/}, all the ratings in the Netflix data for the same user could be revealed. Lastly, in 2014, celebrities in New York were able to be tracked by combining taxi route data released via freedom of information requests, a de-hashing of the taxi license numbers (which were based on md5) and with geo-tagged photos of the celebrities entering/exiting certain taxies \cite{Tockar:2014wa}.

$k$-anonymity was proposed by Sweeney \cite{Sweeney:2002:KAM:774544.774552} to tackle the challenge of linkage attacks on anonymised datasets. Using $k$-anonymity, data is suppressed such that $k-1$ or more individuals possess the same attributes used to create a quasi-identifier. Therefore, an identifiable record in a auxiliary dataset would link to multiple records in the anonymous dataset. However $k$-anonymity cannot defend against linkage attacks where a sensitive attribute is shared among a group of individuals with the same quasi-identifier. $l$-diversity builds on $k$-anonymity to ensure that there is diversity within any group of individuals sharing the same quasi-identifier \cite{Machanavajjhala:2006dd}. $t$-closeness builds on both these methods to preserve the distribution of sensitive attributes among any group of individuals sharing the same quasi-identifier \cite{Li:2007hz}. All the methods suffer however when an adversary possesses some knowledge about the sensitive attribute. Research related to improving $k$-anonymity based methods has mostly been abandoned in the literature in preference of methods that offer more rigorous privacy guarantees (such as \hyperref[sec:differential-privacy]{differntial privacy})

\subsubsection{Encryption}
Anonymisation presents several difficult challenges in order to provide statistics about data without disclosing sensitive information. Encrypting data provides better privacy protection but the ability to perform useful statistical analysis on encrypted data requires specialist methods. Homomorphic encryption \cite{Gentry:2010ct} allows for  processing of data in its encrypted form. Earlier efforts (termed ``Somewhat Homomorphic Encryption'') allowed for simple addition and multiplication operations on encrypted data \cite{Acar:2018bu}, but were shortly followed by Fully Homomorphic Encryption allowing for any arbitrary function to be applied to data in ciphertext form to yield an encrypted result \cite{Gentry:2010ct}. 

Despite the apparent advantages of homomorphic encryption to provide privacy to individuals over their data whilst allowing a third party to perform analytics on it, the computational overhead required to perform such operations is very large \cite{gilad2016cryptonets, 2017arXiv171105189H}. IBM's homomorphic library implementation\footnote{https://github.com/shaih/HElib} runs some 50 million times slower than performing calculations on plaintext data \cite{Rist:2018uq}. Due to this computational overhead, applying homomorphic encryption to training on large-scale machine learning data is currently impractical \cite{duimplementing}. Several projects make use of homomorphic encryption for inference on encrypted private data \cite{2018arXiv181109953C, gilad2016cryptonets}

Secure multi-party computation (SMC) \cite{goldreich1998secure} can also be adopted to compute a function on private data owned by many parties such that no party learns anything about others' data - only the output of the function. Many SMC protocols are based on Shamir's secret sharing \cite{Shamir:1979cr} which splits data into $n$ pieces in such a way that at least $k$ pieces are required to reconstruct the original data ($k-1$ pieces reveal nothing about the original data). For example a value $x$ is shared with multiple servers (as $x_A, x_B...$) via an SMC protocol such that the data can only be reconstructed if the shared pieces on $k$ servers are known \cite{Launchbury:2014gu}. Various protocols exist to compute some function over the data held on the different servers via rounds of communication, however the servers involved are assumed to be trustworthy.

\subsubsection{Differential privacy}
\label{sec:differential-privacy}

Differential privacy provides an elegant and rigorous mathematical measure of the level of privacy afforded by a privacy preserving mechanism. A differentially private privacy preserving mechanism acting on very similar datasets will return statistically indistinguishable results.  More formally: Given some privacy mechanism $M$ that maps inputs from domain $D$ to outputs in range $R$ , it is ``almost'' equally likely (by some multiplicative factor $\epsilon$) for any subset of outputs $S \subseteq R$ to occur, regardless of the presence or absence of a single individual in 2 neighbouring datasets $d$ and $d'$ drawn from $D$ (differing by a single individual) \cite{Dwork:2006dw}

\begin{equation}
    Pr[M(d) \in S] \leq e^{\epsilon}Pr[M(d') \in S].
\end{equation}

Here, $d$ and $d'$ are interchangeable with the same outcome. This privacy guarantee protects individuals from being identified within the dataset as the result from the mechanism should be essentially the same regardless of whether the individual appeared in the original dataset or not. Differential privacy is an example of a perturbative privacy preserving method, as the privacy guarantee is achieved by the addition of noise to the true output. This noise is commonly drawn from a Laplacian distribution \cite{Dwork:2006dw} but can also be drawn from a exponential distribution \cite{Dwork:2014gx} or via the novel staircase mechanism \cite{2015ISTSP...9.1176G} that provides greater utility compared to laplacian noise for the same $\epsilon$. The above description of differential privacy is often termed $\epsilon$-differential privacy or strict differential privacy.

The amount of noise required to satisfy $\epsilon$-differential privacy is governed by $\epsilon$ and the sensitivity of the statistic function $Q$ defined by \cite{Dwork:2006dw}:

\begin{equation}
    \label{eq:l1-sensitivity}
    \Delta Q = max(||Q(d) - Q(d')||_1).
\end{equation}

This maximum is evaluated over all neighbouring datasets in the set $D$ differing by a single individual. The output of the mechanism using noise drawn from the Laplacian distribution is then:

\begin{equation}
    M(d) = Q(d) + Laplace \big(0, \frac{\Delta Q}{\epsilon}\big).
\end{equation}

A relaxed version of differential privacy known as ($\epsilon$, $\delta$)-differential privacy \cite{Kasiviswanathan:uu} provides greater flexibility in designing privacy preserving mechanisms and greater resistance to attacks making use of auxiliary information \cite{Dwork:2014gx}:

\begin{equation}
    Pr[M(d) \in S] \leq e^{\epsilon}Pr[M(d') \in S] + \delta.
\end{equation}

Whereas $\epsilon$-differential privacy provides a privacy guarantee even for results with extremely small probabilities, the $\delta$ in ($\epsilon$, $\delta$)-differential privacy accounts for the small probability that the privacy guarantee of ordinary $\epsilon$-differential privacy is broken.

The Gaussian mechanism is commonly used to add noise to satisfy ($\epsilon$, $\delta$)-differential privacy \cite{Zhu2017}, but instead of the L1 norm used in \autoref{eq:l1-sensitivity}, the noise is scaled to the L2 norm:

\begin{equation}
    \label{eq:l2-sensitivity}
    \Delta_2 Q = max(||Q(d) - Q(d')||_2).
\end{equation}

The following mechanism then satisfies ($\epsilon$, $\delta$)-differential privacy (given $\epsilon, \delta \in (0,1)$):

\begin{equation}
    M(d) = Q(d) + \frac{\Delta_2 Q}{\epsilon} \mathcal{N} \big(0, 2 \ln{(1.25/\delta)}\big).
\end{equation}

$\epsilon$ is additive for multiple queries \cite{Dwork:2014gx} and therefore an $\epsilon$-budget should be designed to protect private data when queried multiple times. Practically, this means that any differential privacy based system must keep track of who queries what and how often to ensure that some predefined $\epsilon$-budget is not surpassed. In a machine learning setting a method of accounting for the accumulated privacy loss over training iterations \cite{Abadi:2016gi} needs to be employed to maintain an $\epsilon$-budget.


Accumulated knowledge as described above is one of the weaknesses of differential privacy to keep sensitive data private \cite{Dwork:2014gx}. Another is collusion. If multiple users collude in the querying of the data (sharing the results of queries with one another) the $\epsilon$-budget for any single user might be breached. Finally, suppose an $\epsilon$-budget is assigned for each individual query; a user making queries on correlated data will use only the budget for each query, yet may be able to gain more information due to the fact that two quantities are correlated (e.g. income and rent). Clearly, large $\epsilon$ (or large $\epsilon$-budgets) introduce greater risk of privacy breaches than small ones but selecting an appropriate $\epsilon$ is a non-trivial issue. Lee and Clinton \cite{Lee:2011cv} discuss the means by which $\epsilon$ might be selected for a given problem but identify that in order to do so, the dataset and the queries on the dataset should be known ahead of time.

Noise addition can be applied in two separate scenarios. Given a trusted data curator, noise can be added to queries on a static dataset, introducing only minimal noise per query. This can be considered as a global privacy setting. Conversely in a local privacy setting, no such trusted curator exits. Local differential privacy applies when noise is added to each sample before collection/aggregation. For example, the randomised response technique \cite{Warner:1965fj} allow participants to answer a question truthfully or randomly based on the flip of a coin. Each participant therefore has plausible deniability for their answer, yet statistics can still be estimated on the population given enough data (and a fair coin flip). Each individual sample is extremely noisy in the local case due to the high vulnerability of a single sample being leaked, however, aggregated in volume, the noise can be filtered out to an extent to reveal population statistics. Federated learning is another example of where local differential privacy is useful \cite{Geyer:2017uk}. Adding noise to the updates during training rounds on local user data prior to aggregation by an untrusted parameter server provides greater privacy to the user and their contributions to a global model (discussed further in \autoref{sec:privacy-preservation-in-federated-learning})

Limited examples of practical applications using differential privacy exist outside of academia. Apple implemented differential privacy in its iOS 10 operating system for the iPhone \cite{Anonymous:yiD6HFFr} in order collect statistics on emoji suggestions and safari crash reports \cite{203940}. Google also collect usage statistics for the Chrome internet browser using differential privacy via multiple rounds of the randomised response technique coupled with a bloom filter to encode the domain names of sites a user has visited \cite{Erlingsson:2014fp}. Both these applications use local differential privacy to protect the individual's privacy but rely on large numbers of participating users in order to determine accurate overall statistics.

Future research and applications of differential privacy are likely to focus on improving utility whilst retaining good privacy guarantees in order for greater adoption by the IT industry.
\section{Privacy preservation in federated learning}
\label{sec:privacy-preservation-in-federated-learning}

Federated learning already increases the level of privacy afforded to an individual over traditional machine learning on a static dataset. It mitigates the storage of sensitive personal data by a third party and prevents a third party from performing learning tasks on the data for which the individual had not initially given permission. Additionally, inference does not require that further sensitive data be sent to a third party as the global model is available to the individual on their own private device \cite{Shokri:2015dr}. Despite these privacy improvements, the weight/gradient updates uploaded by individuals may reveal information about the user's data, especially if certain weights in the weight matrix are sensitive to specific features or values in the individual's data (for example, specific words in a language prediction model \cite{2017arXiv171006963B}). These updates are available to any client participating in federated learning as well as the aggregating server.

\begin{table*}
\caption{A summary of important contributions to federated learning research with a focus on privacy enhancing mechanisms (DP = Differential privacy, HE = Homomorphic encryption, SMC = Secure multi-party computation)}
\label{table:federated-learning-with-PEM-research}
\centering
\small
\begin{tabular}{c p{0.3in} p{2.2in} p{0.4in} p{1.3in}}
\toprule
\textbf{Ref} & \textbf{Year} & \textbf{Major contribution} & \textbf{Privacy mechanism} & \textbf{Privacy details}\\
\midrule
\cite{Shokri:2015dr} & 2015 & Description of a selective distibuted gradient descent method to reduce communication and the application of differential privacy to protect the model parameter updates & DP       & Batch-level DP, $\epsilon$-DP (Laplace mechanism)\\
\midrule
\cite{Abadi:2016gi} & 2016 & Description of an efficient accounting method for accumulating privacy losses while training a DNN with differential privacy & DP & Batch-level DP, ($\epsilon$-$\delta$)-DP (Gaussian mechanism)\\
\midrule
\cite{Bonawitz:2017gu} & 2017 & New method to provide secure multi-party computation specifically tailored towards federated learning                                                                    & SMC      & Secure aggregation protocol evaluates the average gradients of clients only when a sufficient number send updates\\
\midrule
\cite{Geyer:2017uk} & 2017 & Method for providing user-level differential privacy for federated learning with only small loss in model utility                                                        & DP       & User-level DP, ($\epsilon$-$\delta$)-DP (Gaussian mechanism)\\
\midrule
\cite{2017arXiv171006963B} & 2017 & Method for providing user-level differential privacy for federated learning without degrading model utility                                                              & DP       & User-level DP, ($\epsilon$-$\delta$)-DP (Gaussian mechanism)\\
\midrule
\cite{Hitaj:2017cd} & 2017 & Demonstration of an attack method on the global model using a generative adversarial network, effective even against record/batch-level DP                    & DP & Attack tested against record/batch-level DP (implemented using \cite{Shokri:2015dr})\\
\midrule
\cite{Zhang:2017gh} & 2017 & Method for encrypting user updates during distributed training, decryptable only when many clients have participated in the distributed learning objective                   & HE, SMC & Gradient updates are encrypted using homomorphic encryption. Aggregate server obtains average gradient over all workers but can only decrypt this result once a certain number of updates have been aggregated\\
\midrule
\cite{2019arXiv190201046B} & 2019 & Description of a full-scale production-ready federated learning system (focusing on mobile devices)                                                                      & SMC      & Optionally makes use of the Secure aggregation protocol in \cite{Bonawitz:2017gu}\\
\bottomrule
\end{tabular}
\end{table*}

Bonawitz et al. \cite{Bonawitz:2017gu} show that devices participating in federated learning can also act as parties involved in SMC to protect the privacy of all user's updates. In their ``Secure Aggregation'' protocol, the aggregating server only learns about client updates in aggregate. Similarly, the $\propto$MDL protocol described in \cite{Zhang:2017gh} uses SMC but also encrypts the gradients on the client using homomorphic encryption. The summation of the encrypted gradients over all participating clients gives an encrypted global gradient, however this summation result can only be decrypted once a threshold number of clients have shared their gradients. Therefore, again, the server can only learn about client updates in aggregate, preserving the privacy of individual contributions.

Researchers at Google have recently described the high-level design of a production-ready federated learning framework \cite{2019arXiv190201046B} based on Tensorflow\footnote{https://www.tensorflow.org/}. This framework includes Secure Aggregation \cite{Bonawitz:2017gu} as an option during training.

Applying SMC to federated learning suffers from increased communication and greater computational complexity in the aggregation process (both for the client and the server). Additionally, the fully trained model available to clients after the federated learning procedure may still leak sensitive data about specific individuals as described earlier. Adversarial attacks on federated learning models can be mitigated by inspecting and filtering out malicious client updates \cite{2018arXiv180804866F}. However, the Secure Aggregation protocol \cite{Bonawitz:2017gu} prevents the inspection of individual updates and therefore cannot defend against such poisoning attacks \cite{2018arXiv180700459B} in this way.

While SMC achieves privacy through increased computational complexity, differential privacy trades off model utility for increased privacy. Additionally, differential privacy protects individual's contributions to the model during training and once the model is fully trained. Differential privacy has been applied in multiple cases to mitigate the issue of publishing sensitive weight updates during communication rounds in a federated learning setting. Shokri \& Shmatikov \cite{Shokri:2015dr} describe a communication-efficient method for federated learning of a deep learning model tested on the MNIST and SVHN datasets. They select only a fraction of the local gradient updates to share with a central server but also experiment with adding noise to the updates to satisfy differential privacy and protect the contributions of individuals to the global model. An $\epsilon$-budget is divided and spent on selecting gradients above a certain threshold and on publishing the gradients. Judging the sensitivity of SGD is achieved by bounding the gradients between $[-\gamma, \gamma]$ ($\gamma$ is set to some small number). Laplacian noise is generated using this sensitivity and added to the updates prior to selection/publishing. The authors show that their differentially private method outperforms standalone training (training performed by each client on their own data alone) and approaches the performance of SGD on a non-private static dataset given that enough clients participate in each communication round.

Abadi et al. \cite{Abadi:2016gi} apply a differentially private SGD mechanism to train on the MNIST and CIFAR-10 image datasets. They show they can acheive 97\% accuracy on MNIST (1.3\% worse than non-differentially private baseline) and 73\% accuracy on CIFAR-10 (7\% worse than non-differentially private baseline) using a modest neural network and principle component analysis to reduce the dimensionality of the input space. This is achieved using an ($\epsilon$, $\delta$)-differential privacy of (8, $10^{-5}$). The authors also introduce a privacy accountant to monitor the accumulated privacy loss over all training operations based on moments of the privacy loss random variable. The authors point out that the privacy loss is minimal for such a large number of parameters and training examples.

Geyer et al. \cite{Geyer:2017uk} make use of the moments privacy accountant from \cite{Abadi:2016gi} and evaluate the accumulated $\delta$ during training. Once the accumulated $\delta$ reaches a given threshold, training is halted. Intuitively, training is halted once the risk of the privacy guarantee being broken becomes too probable. This method of federated learning protects the privacy of an individual's participation in training over their entire local dataset as opposed to a single data point during training as in \cite{Abadi:2016gi}. The authors show that with a sufficiently large number of clients participating in the federated optimisation, only a minor drop in performance is recorded whilst maintaining a good level of privacy over the individual's data. Similarly, McMahan et al. \cite{2017arXiv171006963B} apply user-level differential privacy (noise is added using sensitivity measured at the user-level rather than sample or mini-batch level) via the moments privacy accountant introduced in \cite{Abadi:2016gi}.

A method for attacking deep learning models trained via federated learning has been proposed in \cite{Hitaj:2017cd}. This approach involves a malicious user participating in federated training whose alternative objective is to train a generative adversarial network (GAN) to generate realistic examples from the globally shared model during training. The authors show that even a model trained with differentially private updates is susceptible to the attack but that it could be defended against with user-level or device-level differential privacy such as that which is described in \cite{Geyer:2017uk} and \cite{2017arXiv171006963B}.

An alternative method to perform machine learning on private data is via a knowledge distillation-like approach. Private Aggregation of Teacher Ensembles (PATE) \cite{Papernot:2016uu} trains a student model (which is published and used for inference) using many teacher models in an ensemble. Neither the sensitive data available to the teacher models, nor the teacher models themselves are ever published. The teacher models once trained on the sensitive data are then used to label public data in a semi-supervised fashion via voting for the predicted class. The votes cast by the teachers have noise generated via a Laplacian distribution added to preserve the privacy of their predictions. This approach requires that public data is available to train the student model, however shows better performance than \cite{Abadi:2016gi} and \cite{Shokri:2015dr} whilst maintaining privacy guarantees (($\epsilon$, $\delta$)-differential privacy of (2.04, $10^{-5}$) and (8.19, $10^{-6}$) on the MINST and SVHN datasets respectively). Further improvements to PATE show that the method can scale to large multi-class problems \cite{Papernot:2018tj}.
\section{Challenges in applying privacy-preserving federated learning to the IoT}
\label{sec:challenges-and-future-directions}
In this section, we identify and outline some promising areas to develop privacy-preserving federated learning research, particularly focused on IoT environments.

\subsection{Optimal model architecture/hyperparameters}
Federated learning precludes seeing the data that a model is trained on. On a traditionally centralised dataset, a deep learning architecture and hyperparameters can be selected via a validation strategy. However to follow the same approach in federated learning to find an optimal architecture or set the optimal hyperparameters to produce good models would require training many models on user devices (possibly incurring unacceptable amounts of battery power and bandwidth). Therefore novel research is required to tackle this specific problem, unique to federated learning.

\subsection{Continual learning}
Training a machine learning model is an expensive and time-consuming task and this can be significantly worse in the federated learning setting. As data distributions evolve over time, a trained model's performance deteriorates. To avoid the cost of federated training many times over, research into methods for improving how a model learns is congruent to the federated learning objective over time. Methods such as meta-learning, online learning and continual learning will be important here which will have specific challenges unique to the distributed nature of federated learning.

\subsection{Better privacy preserving methods}
As seen in this review, there is an observable tradeoff between the performance of a model and the privacy that is afforded to a user. Further research is ongoing into differential privacy accounting methods that introduce less noise into the model (thus improving utility) for the same level of privacy (as judged by the $\epsilon$ parameter). Likewise, further research is required to vastly reduce the computational burden of methods such as homomorphic encryption and secure multi-party computation in order for them to become common-use methods for preserving privacy for large-scale machine learning tasks.

\subsection{Federated learning combined with fog computing}
Reducing the latency between rounds of training in federated learning is desirable to train models quickly. Fog computing nodes could feasibly be leveraged as aggregating servers to remove the round-trip communication between clients and cloud servers in the aggregation step of federated learning. Fog computing could also bring other benefits, such as sharing the computational burden by hierarchically aggregating many large client models.

\subsection{Federated learning on low power devices}
Training deep networks on resource constrained and low power devices poses specific challenges for federated learning. Much of the research into federated learning focusses on mobile devices such as smartphones with abundant compute, storage and power capabilities. As such, new methods are required for reducing the amount of work individual devices need to do to contribute to training (perhaps using the model parallelism approach seen in \cite{Dean:2012wx} or training only certain deep network layers on subsets of devices.)

\section{Conclusion}
\label{sec:conclusion}
Deep learning has shown impressive successes in the fields of computer vision, speech recognition and language modelling. With the exploding increase in deployments of IoT devices, naturally, deep learning is starting to be applied at the edge of the network on mobile and resource-limited embedded devices. This environment however presents difficult challenges for training deep models due to their energy, compute and memory requirements. Beyond this, a model's utility is strictly limited to the data available to the edge device. Allowing machines close to the edge of the network to train on data produced by edge devices (as in fog computing) risks privacy breaches of such data. Federated learning has been shown to be a good solution to improve deep learning models while maintaining the privacy of the raw data.

Federated learning presents a new field of research but has great potential for improving the privacy of training data and giving users control of how their data is used by third parties. Combining federated learning with privacy mechanisms such as differential privacy further secures user data from adversaries with the inclination and means to reverse-engineer parameter updates in distributed SGD procedures. Differential privacy as applied to machine learning is also in its infancy and challenges remain to provide good privacy guarantees whilst simultaneously limiting the required communication costs in a federated setting.

The intersection of federated learning, differential privacy and IoT data represents a fruitful area of research. Performing deep learning efficiently on resource-constrained devices while preserving privacy and utility poses a real challenge. Additionally, the nature of IoT data as opposed to internet data for private federated learning deserves more attention from the research community. IoT data is often represented by highly skewed non-IID data with high temporal variability. This is a challenge that needs to be overcome for federated learning to flourish in edge environments.

\section*{Acknowledgements}
This work is partly supported by the SEND project (grant ref. 32R16P00706) funded by ERDF and BEIS.

\bibliographystyle{IEEEtran}
\bibliography{../main}

\begin{thebibliography}{100}
\providecommand{\url}[1]{#1}
\csname url@samestyle\endcsname
\providecommand{\newblock}{\relax}
\providecommand{\bibinfo}[2]{#2}
\providecommand{\BIBentrySTDinterwordspacing}{\spaceskip=0pt\relax}
\providecommand{\BIBentryALTinterwordstretchfactor}{4}
\providecommand{\BIBentryALTinterwordspacing}{\spaceskip=\fontdimen2\font plus
\BIBentryALTinterwordstretchfactor\fontdimen3\font minus
  \fontdimen4\font\relax}
\providecommand{\BIBforeignlanguage}[2]{{%
\expandafter\ifx\csname l@#1\endcsname\relax
\typeout{** WARNING: IEEEtran.bst: No hyphenation pattern has been}%
\typeout{** loaded for the language `#1'. Using the pattern for}%
\typeout{** the default language instead.}%
\else
\language=\csname l@#1\endcsname
\fi
#2}}
\providecommand{\BIBdecl}{\relax}
\BIBdecl

\bibitem{Gartner:2018wt}
\BIBentryALTinterwordspacing
{Gartner}. (2018, Nov.) {Gartner Identifies Top 10 Strategic IoT Technologies
  and Trends}. [Online]. Available:
  \url{https://www.gartner.com/en/newsroom/press-releases/2018-11-07-gartner-identifies-top-10-strategic-iot-technologies-and-trends}
\BIBentrySTDinterwordspacing

\bibitem{Bonomi:2012kna}
F.~Bonomi, R.~Milito, J.~Zhu, and S.~Addepalli, ``{Fog computing and its role
  in the internet of things},'' in \emph{SIGCOMM 2012 MCC workshop}.\hskip 1em
  plus 0.5em minus 0.4em\relax New York, New York, USA: ACM, Aug. 2012, pp.
  13--16.

\bibitem{Ai:2018kj}
Y.~Ai, M.~Peng, and K.~Zhang, ``{Edge computing technologies for Internet of
  Things: a primer},'' \emph{Digital Communications and Networks}, vol.~4,
  no.~2, pp. 77--86, Apr. 2018.

\bibitem{Bittencourt:2018cr}
L.~Bittencourt, R.~Immich, R.~Sakellariou, N.~Fonseca, E.~Madeira, M.~Curado,
  L.~Villas, L.~DaSilva, C.~Lee, and O.~Rana, ``{The Internet of Things, Fog
  and Cloud continuum: Integration and challenges},'' \emph{Internet of
  Things}, vol. 3{\textendash}4, pp. 134--155, Oct. 2018.

\bibitem{Anonymous:QblbN7U6}
\BIBentryALTinterwordspacing
{OpenFog Consortium}. (2017, Feb.) {OpenFog Reference Architecture for Fog
  Computing }. [Online]. Available:
  \url{https://www.openfogconsortium.org/wp-content/uploads/OpenFog_Reference_Architecture_2_09_17-FINAL.pdf}
\BIBentrySTDinterwordspacing

\bibitem{Bonomi:2014is}
F.~Bonomi, R.~Milito, P.~Natarajan, and J.~Zhu, ``{Fog Computing: A Platform
  for Internet of Things and Analytics},'' in \emph{Big Data and Internet of
  Things: A Roadmap for Smart Environments}.\hskip 1em plus 0.5em minus
  0.4em\relax Cham: Springer, Cham, 2014, pp. 169--186.

\bibitem{Li:2018hz}
H.~Li, K.~Ota, and M.~Dong, ``{Learning IoT in Edge: Deep Learning for the
  Internet of Things with Edge Computing},'' \emph{IEEE Network}, vol.~32,
  no.~1, pp. 96--101, 2018.

\bibitem{BenNun:2019if}
T.~Ben-Nun and T.~Hoefler, ``{Demystifying Parallel and Distributed Deep
  Learning},'' \emph{ACM Computing Surveys (CSUR)}, vol.~52, no.~4, pp. 1--43,
  Sep. 2019.

\bibitem{mcmahan2017communication}
B.~McMahan, E.~Moore, D.~Ramage, S.~Hampson, and B.~A. y~Arcas,
  ``{Communication-Efficient Learning of Deep Networks from Decentralized
  Data},'' in \emph{Artificial Intelligence and Statistics}, 2017, pp.
  1273--1282.

\bibitem{Abadi:2016gi}
M.~Abadi, A.~Chu, I.~Goodfellow, H.~B. McMahan, I.~Mironov, K.~Talwar, and
  L.~Zhang, ``{Deep Learning with Differential Privacy},'' in \emph{the 2016
  ACM SIGSAC Conference}.\hskip 1em plus 0.5em minus 0.4em\relax ACM, Oct.
  2016, pp. 308--318.

\bibitem{Shokri:2015dr}
R.~Shokri and V.~Shmatikov, ``{Privacy-Preserving Deep Learning},'' in
  \emph{the 22nd ACM SIGSAC Conference}.\hskip 1em plus 0.5em minus 0.4em\relax
  ACM, Oct. 2015, pp. 1310--1321.

\bibitem{Wang:2019kq}
S.~Wang, T.~Tuor, T.~Salonidis, K.~K. Leung, C.~Makaya, T.~He, and K.~Chan,
  ``{Adaptive Federated Learning in Resource Constrained Edge Computing
  Systems},'' \emph{IEEE Journal on Selected Areas in Communications}, vol.~37,
  no.~6, pp. 1205--1221, Jun. 2019.

\bibitem{Dean:2012wx}
J.~Dean, G.~S. Corrado, R.~Monga, K.~Chen, M.~Devin, Q.~V. Le, M.~Z. Mao,
  M.~Ranzato, A.~Senior, P.~Tucker, K.~Yang, and A.~Y. Ng, ``{Large scale
  distributed deep networks},'' in \emph{Advances in neural information
  processing systems}.\hskip 1em plus 0.5em minus 0.4em\relax Curran Associates
  Inc., Dec. 2012, pp. 1223--1231.

\bibitem{Li:tt}
M.~Li, D.~G. Andersen, J.~W. Park, A.~J. Smola, and A.~Ahmed, ``{Scaling
  Distributed Machine Learning with the Parameter Server.}'' \emph{OSDI},
  vol.~14, pp. 583--598, 2014.

\bibitem{2015Natur.521..436L}
Y.~Lecun, Y.~Bengio, and G.~E. Hinton, ``{Deep learning},'' \emph{Nature}, vol.
  521, no.~7, pp. 436--444, May 2015.

\bibitem{Goodfellow-et-al-2016}
I.~Goodfellow, Y.~Bengio, and A.~Courville, \emph{{Deep Learning}},
  1st~ed.\hskip 1em plus 0.5em minus 0.4em\relax Cambridge, Massachusetts: MIT
  Press, 2016.

\bibitem{RUMELHART:1986to}
D.~E. Rumelhart, G.~E. Hinton, and R.~J. Williams, ``{Learning internal
  representations by error propagation},'' in \emph{Parallel distributed
  processing explorations in the microstructure of cognition}.\hskip 1em plus
  0.5em minus 0.4em\relax Cambridge, Massachusetts: MIT Press, Jan. 1986, pp.
  318--362.

\bibitem{Krizhevsky:2017gx}
A.~Krizhevsky, I.~Sutskever, and G.~E. Hinton, ``{ImageNet Classification with
  Deep Convolutional Neural Networks},'' \emph{Communications of the Acm},
  vol.~60, no.~6, pp. 84--90, Jun. 2017.

\bibitem{Chilimbi:vp}
T.~M. Chilimbi, Y.~Suzue, J.~Apacible, and K.~Kalyanaraman, ``{Project Adam:
  Building an Efficient and Scalable Deep Learning Training System.}'' in
  \emph{OSDI}, 2014, pp. 571--582.

\bibitem{Recht:2011wo}
B.~Recht, C.~R{\'e}, S.~Wright, and F.~Niu, ``{Hogwild: A Lock-Free Approach to
  Parallelizing Stochastic Gradient Descent},'' in \emph{Advances in neural
  information processing systems}, 2011, pp. 693--701.

\bibitem{Ho:2013wd}
Q.~Ho, J.~Cipar, H.~Cui, S.~Lee, J.~K. Kim, P.~B. Gibbons, G.~A. Gibson,
  G.~Ganger, and E.~P. Xing, ``{More Effective Distributed ML via a Stale
  Synchronous Parallel Parameter Server},'' in \emph{Advances in neural
  information processing systems}, 2013, pp. 1223--1231.

\bibitem{2016arXiv160104033O}
A.~Odena, ``{Faster Asynchronous SGD},'' \emph{arXiv.org}, p. arXiv:1601.04033,
  Jan. 2016.

\bibitem{2015arXiv151105950Z}
W.~Zhang, S.~Gupta, X.~Lian, and J.~Liu, ``{Staleness-aware Async-SGD for
  Distributed Deep Learning},'' in \emph{Proceedings of the Twenty-Fifth
  International Joint Conference on Artificial Intelligence}, Nov. 2015, pp.
  2350--2356.

\bibitem{Hsieh:tv}
K.~Hsieh, A.~Harlap, N.~Vijaykumar, D.~Konomis, G.~R. Ganger, and P.~B.
  Gibbons, ``{Gaia: Geo-Distributed Machine Learning Approaching LAN Speeds.}''
  in \emph{NSDI}, 2017, pp. 629--647.

\bibitem{Jiang:2017cs}
J.~Jiang, B.~Cui, C.~Zhang, and L.~Yu, ``{Heterogeneity-aware Distributed
  Parameter Servers},'' in \emph{the 2017 ACM International Conference}.\hskip
  1em plus 0.5em minus 0.4em\relax New York, New York, USA: ACM Press, 2017,
  pp. 463--478.

\bibitem{2018arXiv180305880D}
J.~Daily, A.~Vishnu, C.~Siegel, T.~Warfel, and V.~Amatya, ``{GossipGraD:
  Scalable Deep Learning using Gossip Communication based Asynchronous Gradient
  Descent},'' \emph{arXiv.org}, p. arXiv:1803.05880, Mar. 2018.

\bibitem{2016arXiv161104581J}
P.~H. Jin, Q.~Yuan, F.~Iandola, and K.~Keutzer, ``{How to scale distributed
  deep learning?}'' \emph{arXiv.org}, p. arXiv:1611.04581, Nov. 2016.

\bibitem{SundharRam:2009fa}
S.~Sundhar~Ram, A.~Nedic, and V.~V. Veeravalli, ``{Asynchronous gossip
  algorithms for stochastic optimization},'' in \emph{2009 International
  Conference on Game Theory for Networks (GameNets)}.\hskip 1em plus 0.5em
  minus 0.4em\relax IEEE, 2009, pp. 80--81.

\bibitem{Ba:2014tr}
J.~Ba and R.~Caruana, ``{Do Deep Nets Really Need to be Deep?}'' in
  \emph{Advances in neural information processing systems}, 2014, pp.
  2654--2662.

\bibitem{Chebotar:2016bi}
Y.~Chebotar and A.~Waters, ``{Distilling Knowledge from Ensembles of Neural
  Networks for Speech Recognition},'' in \emph{Interspeech 2016}.\hskip 1em
  plus 0.5em minus 0.4em\relax ISCA, Sep. 2016, pp. 3439--3443.

\bibitem{2015arXiv150302531H}
G.~E. Hinton, O.~Vinyals, and J.~Dean, ``{Distilling the Knowledge in a Neural
  Network},'' \emph{arXiv.org}, p. arXiv:1503.02531, Mar. 2015.

\bibitem{Liang:vq}
Y.~Liang, M.~F. Balcan, and V.~Kanchanapally, ``{Distributed PCA and k-means
  clustering},'' in \emph{The Big Learning Workshop at NIPS}, 2013.

\bibitem{2016arXiv161002527K}
J.~Kone{\v c}n{\'{y}}, H.~B. McMahan, D.~Ramage, and P.~Richt{\'a}rik,
  ``{Federated Optimization: Distributed Machine Learning for On-Device
  Intelligence},'' \emph{arXiv.org}, p. arXiv:1610.02527, Oct. 2016.

\bibitem{Johnson:2013:ASG:2999611.2999647}
R.~Johnson and T.~Zhang, ``{Accelerating Stochastic Gradient Descent Using
  Predictive Variance Reduction},'' in \emph{Proceedings of the 26th
  International Conference on Neural Information Processing Systems - Volume
  1}.\hskip 1em plus 0.5em minus 0.4em\relax USA: Curran Associates Inc., 2013,
  pp. 315--323.

\bibitem{2018arXiv180600582Z}
Y.~Zhao, M.~Li, L.~Lai, N.~Suda, D.~Civin, and V.~Chandra, ``{Federated
  Learning with Non-IID Data},'' \emph{arXiv.org}, p. arXiv:1806.00582, Jun.
  2018.

\bibitem{mlsys2020_176}
I.~Dhillon, D.~Papailiopoulos, and V.~Sze, Eds., \emph{{Federated Optimization
  in Heterogeneous Networks}}, 2020.

\bibitem{karimireddy2019scaffold}
S.~P. Karimireddy, S.~Kale, M.~Mohri, S.~J. Reddi, S.~U. Stich, and A.~T.
  Suresh, ``{Scaffold: Stochastic controlled averaging for federated
  learning},'' \emph{arXiv preprint arXiv:1910.06378}, 2020.

\bibitem{Konecny:2016ts}
J.~Kone{\v c}n{\'{y}}, H.~B. McMahan, F.~X. Yu, P.~Richt{\'a}rik, A.~T. Suresh,
  and D.~Bacon, ``{Federated Learning: Strategies for Improving Communication
  Efficiency},'' \emph{arXiv.org}, p. arXiv:1610.05492, Oct. 2016.

\bibitem{2017arXiv170510467S}
V.~Smith, C.-K. Chiang, M.~Sanjabi, and A.~Talwalkar, ``{Federated Multi-Task
  Learning},'' in \emph{Advances in neural information processing systems}, May
  2017, p. arXiv:1705.10467.

\bibitem{2018arXiv180700459B}
E.~Bagdasaryan, A.~Veit, Y.~Hua, D.~Estrin, and V.~Shmatikov, ``{How To
  Backdoor Federated Learning},'' \emph{arXiv.org}, p. arXiv:1807.00459, Jul.
  2018.

\bibitem{2018arXiv180804866F}
C.~Fung, C.~J.~M. Yoon, and I.~Beschastnikh, ``{Mitigating Sybils in Federated
  Learning Poisoning},'' \emph{arXiv.org}, p. arXiv:1808.04866, Aug. 2018.

\bibitem{Hard:2018tda}
A.~Hard, K.~Rao, R.~Mathews, F.~Beaufays, S.~Augenstein, H.~Eichner, C.~Kiddon,
  and D.~Ramage, ``{Federated Learning for Mobile Keyboard Prediction},''
  \emph{arXiv.org}, Nov. 2018.

\bibitem{2018arXiv181005512L}
D.~Leroy, A.~Coucke, T.~Lavril, T.~Gisselbrecht, and J.~Dureau, ``{Federated
  Learning for Keyword Spotting},'' \emph{arXiv.org}, p. arXiv:1810.05512, Oct.
  2018.

\bibitem{Liu:2018tv}
D.~Liu, T.~Miller, R.~Sayeed, and K.~D. Mandl, ``{FADL: Federated-Autonomous
  Deep Learning for Distributed Electronic Health Record},'' \emph{arXiv.org},
  p. arXiv:1811.11400, Nov. 2018.

\bibitem{2018arXiv180408333N}
T.~Nishio and R.~Yonetani, ``{Client Selection for Federated Learning with
  Heterogeneous Resources in Mobile Edge},'' \emph{arXiv.org}, p.
  arXiv:1804.08333, Apr. 2018.

\bibitem{2019arXiv190410120E}
H.~Eichner, T.~Koren, H.~B. McMahan, N.~Srebro, and K.~Talwar, ``{Semi-Cyclic
  Stochastic Gradient Descent},'' in \emph{International Conference on Machine
  Learning}, Apr. 2019, p. arXiv:1904.10120.

\bibitem{Sattler:2019tb}
F.~Sattler, S.~Wiedemann, K.-R. M{\"u}ller, and W.~Samek, ``{Robust and
  Communication-Efficient Federated Learning from Non-IID Data},''
  \emph{arXiv.org}, p. arXiv:1903.02891, Mar. 2019.

\bibitem{2020arXiv200411791B}
C.~Briggs, Z.~Fan, and P.~Andras, ``{Federated learning with hierarchical
  clustering of local updates to improve training on non-IID data},''
  \emph{arXiv.org}, p. arXiv:2004.11791, Apr. 2020.

\bibitem{Liu:2020hr}
B.~Liu, L.~Wang, M.~Liu, and C.-Z. Xu, ``{Federated Imitation Learning: A Novel
  Framework for Cloud Robotic Systems With Heterogeneous Sensor Data},''
  \emph{IEEE Robotics and Automation Letters}, vol.~5, no.~2, pp. 3509--3516,
  Apr. 2020.

\bibitem{Zhou:2018ez}
W.~Zhou, Y.~Li, S.~Chen, and B.~Ding, ``{Real-Time Data Processing Architecture
  for Multi-Robots Based on Differential Federated Learning},'' in \emph{2018
  IEEE SmartWorld, Ubiquitous Intelligence {\&} Computing, Advanced {\&}
  Trusted Computing, Scalable Computing {\&} Communications, Cloud {\&} Big
  Data Computing, Internet of People and Smart City Innovation
  (SmartWorld/SCALCOM/UIC/ATC/CBDCom/IOP/SCI)}.\hskip 1em plus 0.5em minus
  0.4em\relax IEEE, 2018, pp. 462--471.

\bibitem{Fantacci:2020db}
R.~Fantacci and B.~Picano, ``{Federated learning framework for mobile edge
  computing networks},'' \emph{CAAI Transactions on Intelligence Technology},
  vol.~5, no.~1, pp. 15--21, Mar. 2020.

\bibitem{Yu:2018fx}
Z.~Yu, J.~Hu, G.~Min, H.~Lu, Z.~Zhao, H.~Wang, and N.~Georgalas, ``{Federated
  Learning Based Proactive Content Caching in Edge Computing},'' in
  \emph{GLOBECOM 2018 - 2018 IEEE Global Communications Conference}.\hskip 1em
  plus 0.5em minus 0.4em\relax IEEE, 2018, pp. 1--6.

\bibitem{Lu:2019jf}
Y.~Lu, X.~Huang, Y.~Dai, S.~Maharjan, and Y.~Zhang, ``{Differentially Private
  Asynchronous Federated Learning for Mobile Edge Computing in Urban
  Informatics},'' \emph{IEEE Transactions on Industrial Informatics}, vol.~16,
  no.~3, pp. 2134--2143, 2019.

\bibitem{Ye:2020em}
D.~Ye, R.~Yu, M.~Pan, and Z.~Han, ``{Federated Learning in Vehicular Edge
  Computing: A Selective Model Aggregation Approach},'' \emph{IEEE Access},
  vol.~8, pp. 23\,920--23\,935, 2020.

\bibitem{Saputra:2019ha}
Y.~M. Saputra, D.~T. Hoang, D.~N. Nguyen, E.~Dutkiewicz, M.~D. Mueck, and
  S.~Srikanteswara, ``{Energy Demand Prediction with Federated Learning for
  Electric Vehicle Networks},'' in \emph{GLOBECOM 2019 - 2019 IEEE Global
  Communications Conference}.\hskip 1em plus 0.5em minus 0.4em\relax IEEE,
  2019, pp. 1--6.

\bibitem{Nguyen:2019by}
T.~D. Nguyen, S.~Marchal, M.~Miettinen, H.~Fereidooni, N.~Asokan, and A.-R.
  Sadeghi, ``{D{\"I}oT: A Federated Self-learning Anomaly Detection System for
  IoT},'' in \emph{2019 IEEE 39th International Conference on Distributed
  Computing Systems (ICDCS)}.\hskip 1em plus 0.5em minus 0.4em\relax IEEE,
  2019, pp. 756--767.

\bibitem{Mowla:2020fx}
N.~I. Mowla, N.~H. Tran, I.~Doh, and K.~Chae, ``{Federated Learning-Based
  Cognitive Detection of Jamming Attack in Flying Ad-Hoc Network},'' \emph{IEEE
  Access}, vol.~8, pp. 4338--4350, 2020.

\bibitem{Chen:2020dj}
M.~Chen, O.~Semiari, W.~Saad, X.~Liu, and C.~Yin, ``{Federated Echo State
  Learning for Minimizing Breaks in Presence in Wireless Virtual Reality
  Networks},'' \emph{Ieee Transactions on Wireless Communications}, vol.~19,
  no.~1, pp. 177--191, Jan. 2020.

\bibitem{Sozinov:2018fk}
K.~Sozinov, V.~Vlassov, and S.~Girdzijauskas, ``{Human Activity Recognition
  Using Federated Learning},'' in \emph{2018 IEEE Intl Conf on Parallel {\&}
  Distributed Processing with Applications, Ubiquitous Computing {\&}
  Communications, Big Data {\&} Cloud Computing, Social Computing {\&}
  Networking, Sustainable Computing {\&} Communications
  (ISPA/IUCC/BDCloud/SocialCom/SustainCom)}.\hskip 1em plus 0.5em minus
  0.4em\relax IEEE, 2018, pp. 1103--1111.

\bibitem{Miotto:2017km}
R.~Miotto, F.~Wang, S.~Wang, X.~Jiang, and J.~T. Dudley, ``{Deep learning for
  healthcare: review, opportunities and challenges},'' \emph{Briefings in
  Bioinformatics}, vol.~19, no.~6, pp. 1236--1246, May 2017.

\bibitem{Bhagoji:2019vz}
\emph{{Analyzing Federated Learning through an Adversarial Lens}}.\hskip 1em
  plus 0.5em minus 0.4em\relax PMLR, May 2019.

\bibitem{2015arXiv151000149H}
S.~Han, H.~Mao, and W.~J. Dally, ``{Deep Compression: Compressing Deep Neural
  Networks with Pruning, Trained Quantization and Huffman Coding},''
  \emph{arXiv.org}, p. arXiv:1510.00149, Oct. 2015.

\bibitem{Han:2015vz}
S.~Han, J.~Pool, J.~Tran, and W.~Dally, ``{Learning both Weights and
  Connections for Efficient Neural Network},'' in \emph{Advances in neural
  information processing systems}, 2015, pp. 1135--1143.

\bibitem{2015arXiv150202551G}
S.~Gupta, A.~Agrawal, K.~Gopalakrishnan, and P.~Narayanan, ``{Deep Learning
  with Limited Numerical Precision},'' \emph{arXiv.org}, p. arXiv:1502.02551,
  Feb. 2015.

\bibitem{2015arXiv151104561D}
T.~Dettmers, ``{8-Bit Approximations for Parallelism in Deep Learning},''
  \emph{arXiv.org}, p. arXiv:1511.04561, Nov. 2015.

\bibitem{Seide:vu}
F.~Seide, H.~Fu, J.~Droppo, G.~Li, and D.~Yu, ``{1-bit stochastic gradient
  descent and its application to data-parallel distributed training of speech
  dnns},'' in \emph{Fifteenth Annual Conference of the International Speech
  Communication Association}, 2014.

\bibitem{Hardy:hu}
C.~Hardy, E.~Le~Merrer, and B.~Sericola, ``{Distributed deep learning on
  edge-devices: Feasibility via adaptive compression},'' in \emph{2017 IEEE
  16th International Symposium on Network Computing and Applications
  (NCA)}.\hskip 1em plus 0.5em minus 0.4em\relax IEEE, pp. 1--8.

\bibitem{2017arXiv171201887L}
Y.~Lin, S.~Han, H.~Mao, Y.~Wang, and W.~J. Dally, ``{Deep Gradient Compression:
  Reducing the Communication Bandwidth for Distributed Training},''
  \emph{arXiv.org}, p. arXiv:1712.01887, Dec. 2017.

\bibitem{Gandomi:2015hh}
A.~Gandomi and M.~Haider, ``{Beyond the hype: Big data concepts, methods, and
  analytics},'' \emph{International Journal of Information Management},
  vol.~35, no.~2, pp. 137--144, Apr. 2015.

\bibitem{Anonymous:2015uv}
{UN General Assembly}, ``{Universal Declaration of Human Rights},'' Oct. 2015.

\bibitem{Anonymous:1966uz}
------, ``{International Covenant on Civil and Political Rights},'' Dec. 1966.

\bibitem{Anonymous:1950wd}
{Council of Europe}, ``{European convention for the protection of human rights
  and fundamental freedoms },'' Nov. 1950.

\bibitem{eu:gdpr}
{European Commision}, ``{Regulation (EU) 2016/679 of the European Parliament
  and of the Council of 27 April 2016 on the protection of natural persons with
  regard to the processing of personal data and on the free movement of such
  data, and repealing Directive 95/46/EC (General Data Protection
  Regulation)},'' \emph{Official Journal of the European Union}, vol. L119, pp.
  1--88, May 2016.

\bibitem{Dwork:2006dw}
C.~Dwork, ``{Differential Privacy},'' in \emph{Automata, Languages and
  Programming}.\hskip 1em plus 0.5em minus 0.4em\relax Berlin, Heidelberg:
  Springer Berlin Heidelberg, 2006, pp. 1--12.

\bibitem{Fung:2010ena}
B.~C.~M. Fung, K.~Wang, R.~Chen, and P.~S. Yu, ``{Privacy-preserving data
  publishing: A survey of recent developments},'' \emph{ACM Computing Surveys
  (CSUR)}, vol.~42, no.~4, pp. 14--53, Jun. 2010.

\bibitem{Greely:2007jc}
H.~T. Greely, ``{The uneasy ethical and legal underpinnings of large-scale
  genomic biobanks},'' \emph{Annual Review of Genomics and Human Genetics},
  vol.~8, no.~1, pp. 343--364, 2007.

\bibitem{Narayanan:2008iu}
A.~Narayanan and V.~Shmatikov, ``{Robust De-anonymization of Large Sparse
  Datasets},'' in \emph{2008 IEEE Symposium on Security and Privacy (sp
  2008)}.\hskip 1em plus 0.5em minus 0.4em\relax IEEE, 2008, pp. 111--125.

\bibitem{Tockar:2014wa}
\BIBentryALTinterwordspacing
A.~Tockar. (2014, Sep.) {Riding with the Stars: Passenger Privacy in the NYC
  Taxicab Dataset}. [Online]. Available:
  \url{https://research.neustar.biz/2014/09/15/riding-with-the-stars-passenger-privacy-in-the-nyc-taxicab-dataset/}
\BIBentrySTDinterwordspacing

\bibitem{Sweeney:2002:KAM:774544.774552}
L.~Sweeney, ``{K-anonymity: A Model for Protecting Privacy},'' \emph{Int. J.
  Uncertain. Fuzziness Knowl.-Based Syst.}, vol.~10, no.~5, pp. 557--570, Oct.
  2002.

\bibitem{Machanavajjhala:2006dd}
A.~Machanavajjhala, J.~Gehrke, D.~Kifer, and M.~Venkitasubramaniam,
  ``{L-diversity: privacy beyond k-anonymity},'' in \emph{22nd International
  Conference on Data Engineering}.\hskip 1em plus 0.5em minus 0.4em\relax IEEE,
  2006, pp. 24--24.

\bibitem{Li:2007hz}
N.~Li, T.~Li, and S.~Venkatasubramanian, ``{t-Closeness: Privacy Beyond
  k-Anonymity and l-Diversity},'' in \emph{2007 IEEE 23rd International
  Conference on Data Engineering}.\hskip 1em plus 0.5em minus 0.4em\relax IEEE,
  2007, pp. 106--115.

\bibitem{Gentry:2010ct}
C.~Gentry, ``{Computing arbitrary functions of encrypted data},''
  \emph{Communications of the Acm}, vol.~53, no.~3, pp. 97--105, Mar. 2010.

\bibitem{Acar:2018bu}
A.~Acar, H.~Aksu, A.~S. Uluagac, and M.~Conti, ``{A Survey on Homomorphic
  Encryption Schemes: Theory and Implementation},'' \emph{ACM Computing Surveys
  (CSUR)}, vol.~51, no.~4, pp. 79--35, Sep. 2018.

\bibitem{gilad2016cryptonets}
R.~Gilad-Bachrach, N.~Dowlin, K.~Laine, K.~Lauter, M.~Naehrig, and J.~Wernsing,
  ``{Cryptonets: Applying neural networks to encrypted data with high
  throughput and accuracy},'' in \emph{International Conference on Machine
  Learning}, 2016, pp. 201--210.

\bibitem{2017arXiv171105189H}
E.~Hesamifard, H.~Takabi, and M.~Ghasemi, ``{CryptoDL: Deep Neural Networks
  over Encrypted Data},'' \emph{arXiv.org}, p. arXiv:1711.05189, Nov. 2017.

\bibitem{Rist:2018uq}
\BIBentryALTinterwordspacing
L.~Rist. (2018, Jan.) {Encrypt your Machine Learning}. [Online]. Available:
  \url{https://medium.com/corti-ai/encrypt-your-machine-learning-12b113c879d6}
\BIBentrySTDinterwordspacing

\bibitem{duimplementing}
Y.~Du, L.~Gustafson, D.~Huang, and K.~Peterson, \emph{{Implementing ML
  Algorithms with HE}}.\hskip 1em plus 0.5em minus 0.4em\relax MIT Course
  6.857: Computer and Network Security, 2017.

\bibitem{2018arXiv181109953C}
E.~Chou, J.~Beal, D.~Levy, S.~Yeung, A.~Haque, and L.~Fei-Fei, ``{Faster
  CryptoNets: Leveraging Sparsity for Real-World Encrypted Inference},''
  \emph{arXiv.org}, p. arXiv:1811.09953, Nov. 2018.

\bibitem{goldreich1998secure}
O.~Goldreich, ``{Secure multi-party computation},'' \emph{Manuscript.
  Preliminary version}, vol.~78, 1998.

\bibitem{Shamir:1979cr}
A.~Shamir, ``{How to share a secret},'' \emph{Communications of the Acm},
  vol.~22, no.~11, pp. 612--613, Nov. 1979.

\bibitem{Launchbury:2014gu}
J.~Launchbury, D.~Archer, T.~DuBuisson, and E.~Mertens, ``{Application-Scale
  Secure Multiparty Computation},'' in \emph{Programming Languages and
  Systems}.\hskip 1em plus 0.5em minus 0.4em\relax Berlin, Heidelberg:
  Springer, Berlin, Heidelberg, Apr. 2014, pp. 8--26.

\bibitem{Dwork:2014gx}
C.~Dwork and A.~Roth, ``{The Algorithmic Foundations of Differential
  Privacy},'' \emph{Foundations and Trends in Theoretical Computer Science},
  vol.~9, no. 3{\textendash}4, pp. 211--407, Aug. 2014.

\bibitem{2015ISTSP...9.1176G}
Q.~Geng, P.~Kairouz, S.~Oh, and P.~Viswanath, ``{The Staircase Mechanism in
  Differential Privacy},'' \emph{IEEE Journal of Selected Topics in Signal
  Processing}, vol.~9, no.~7, pp. 1176--1184, Oct. 2015.

\bibitem{Kasiviswanathan:uu}
S.~P. Kasiviswanathan and A.~Smith, ``{On the'semantics' of differential
  privacy: A bayesian formulation},'' \emph{Journal of Privacy and
  Confidentiality}, vol.~6, no.~1, 2014.

\bibitem{Zhu2017}
T.~Zhu, G.~Li, W.~Zhou, and P.~S. Yu, ``{Preliminary of Differential
  Privacy},'' in \emph{Differential Privacy and Applications}.\hskip 1em plus
  0.5em minus 0.4em\relax Cham: Springer International Publishing, 2017, pp.
  7--16.

\bibitem{Lee:2011cv}
J.~Lee and C.~Clifton, ``{How Much Is Enough? Choosing $\varepsilon$ for
  Differential Privacy},'' in \emph{Information Security}.\hskip 1em plus 0.5em
  minus 0.4em\relax Berlin, Heidelberg: Springer, Berlin, Heidelberg, Oct.
  2011, pp. 325--340.

\bibitem{Warner:1965fj}
S.~L. Warner, ``{Randomized Response: A Survey Technique for Eliminating
  Evasive Answer Bias},'' \emph{Journal of the American Statistical
  Association}, vol.~60, no. 309, p.~63, Mar. 1965.

\bibitem{Geyer:2017uk}
R.~C. Geyer, T.~Klein, and M.~Nabi, ``{Differentially Private Federated
  Learning: A Client Level Perspective},'' \emph{arXiv.org}, p.
  arXiv:1712.07557, Dec. 2017.

\bibitem{Anonymous:yiD6HFFr}
\BIBentryALTinterwordspacing
{Apple}. (2017, Nov.) {Apple Differential Privacy Technical Overview}.
  [Online]. Available:
  \url{https://www.apple.com/privacy/docs/Differential_Privacy_Overview.pdf}
\BIBentrySTDinterwordspacing

\bibitem{203940}
A.~G. Thakurta, ``{Differential Privacy: From Theory to Deployment},'' in
  \emph{USENIX Association}.\hskip 1em plus 0.5em minus 0.4em\relax Vancouver,
  BC: USENIX Association, 2017.

\bibitem{Erlingsson:2014fp}
{\'U}.~Erlingsson, V.~Pihur, and A.~Korolova, ``{RAPPOR: Randomized
  Aggregatable Privacy-Preserving Ordinal Response},'' in \emph{Proceedings of
  the ACM SIGSAC Conference on Computer and Communications Security}.\hskip 1em
  plus 0.5em minus 0.4em\relax ACM, Nov. 2014, pp. 1054--1067.

\bibitem{2017arXiv171006963B}
H.~B. McMahan, D.~Ramage, K.~Talwar, and L.~Zhang, ``{Learning Differentially
  Private Recurrent Language Models},'' \emph{arXiv.org}, p. arXiv:1710.06963,
  Oct. 2017.

\bibitem{Bonawitz:2017gu}
K.~Bonawitz, V.~Ivanov, B.~Kreuter, A.~Marcedone, H.~B. McMahan, S.~Patel,
  D.~Ramage, A.~Segal, and K.~Seth, ``{Practical Secure Aggregation for
  Privacy-Preserving Machine Learning},'' in \emph{Proceedings of the ACM
  SIGSAC Conference on Computer and Communications Security}.\hskip 1em plus
  0.5em minus 0.4em\relax ACM, Oct. 2017, pp. 1175--1191.

\bibitem{Hitaj:2017cd}
B.~Hitaj, G.~Ateniese, and F.~Perez-Cruz, ``{Deep Models Under the GAN},'' in
  \emph{the 2017 ACM SIGSAC Conference}.\hskip 1em plus 0.5em minus 0.4em\relax
  New York, New York, USA: ACM Press, 2017, pp. 603--618.

\bibitem{Zhang:2017gh}
X.~Zhang, S.~Ji, H.~Wang, and T.~Wang, ``{Private, Yet Practical, Multiparty
  Deep Learning},'' in \emph{2017 IEEE 37th International Conference on
  Distributed Computing Systems (ICDCS)}.\hskip 1em plus 0.5em minus
  0.4em\relax IEEE, 2017, pp. 1442--1452.

\bibitem{2019arXiv190201046B}
K.~Bonawitz, H.~Eichner, W.~Grieskamp, D.~Huba, A.~Ingerman, V.~Ivanov,
  C.~Kiddon, J.~Kone{\v c}n{\'{y}}, S.~Mazzocchi, H.~B. McMahan,
  T.~Van~Overveldt, D.~Petrou, D.~Ramage, and J.~Roselander, ``{Towards
  Federated Learning at Scale: System Design},'' \emph{arXiv.org}, p.
  arXiv:1902.01046, Feb. 2019.

\bibitem{Papernot:2016uu}
N.~Papernot, M.~Abadi, {\'U}.~Erlingsson, I.~Goodfellow, and K.~Talwar,
  ``{Semi-supervised Knowledge Transfer for Deep Learning from Private Training
  Data},'' \emph{arXiv.org}, p. arXiv:1610.05755, Oct. 2016.

\bibitem{Papernot:2018tj}
N.~Papernot, S.~Song, I.~Mironov, A.~Raghunathan, K.~Talwar, and
  {\'U}.~Erlingsson, ``{Scalable Private Learning with PATE},''
  \emph{arXiv.org}, p. arXiv:1802.08908, Feb. 2018.

\end{thebibliography}
\end{document}